\documentclass[letterpaper, 10 pt, conference]{ieeeconf}
\usepackage{amsmath,amsfonts}
\usepackage{array}
\usepackage{enumerate}
\usepackage[utf8]{inputenc}
\usepackage[T1]{fontenc}
\usepackage{amsfonts}
\usepackage{amssymb}
\IEEEoverridecommandlockouts 
\usepackage{algorithm}  
\usepackage{algpseudocode}  
\usepackage{amsmath}  
\usepackage{breqn}
\usepackage[caption=false]{subfig}
\usepackage{textcomp}
\usepackage{stfloats}
\usepackage{url}
\usepackage{verbatim}
\usepackage{graphicx}
\usepackage{cite}
\usepackage{diagbox}
\usepackage{amsfonts}
\usepackage{makecell}
\usepackage{adjustbox}
\usepackage{color} 

\usepackage{hyperref}

\title{\LARGE \bf
Hyper-STTN: Hypergraph Augmented Spatial-Temporal Transformer Network for Trajectory Prediction

}

\author{
Weizheng Wang$^{1*}$, 
Baijian Yang$^{1}$, 
Sungeun Hong$^{3}$, 
Wenhai Sun$^{1}$, 
and Byung-Cheol Min$^{1,2}$
\thanks{$^{1}$School of Applied and Creative Computing, Purdue University, West Lafayette, IN, USA. {\tt\small wang5716@purdue.edu, byang@purdue.edu, sun841@purdue.edu}, $^{2}$Department of Computer Science and Department of Intelligent Systems Engineering, Indiana University Bloomington, Bloomington, IN, USA. {\tt\small minb@iu.edu}, $^{3}$Department of Applied Artificial Intelligence, Sungkyunkwan University, Suwon, South Korea. {\tt\small csehong@skku.edu}.}
\thanks{$^{*}$Correspondence to: {\tt\small wang5716@purdue.edu}.}
\thanks{Project website: \url{https://sites.google.com/view/hypersttn}}
}

\begin{document}

\maketitle

\begin{abstract}
Predicting crowd intentions and trajectories is critical for a range of real-world applications, involving social robotics and autonomous driving. Accurately modeling such behavior remains challenging due to the complexity of pairwise spatial-temporal interactions and the heterogeneous influence of groupwise dynamics. To address these challenges, we propose Hyper-STTN, a Hypergraph-based Spatial-Temporal Transformer Network for crowd trajectory prediction. Hyper-STTN constructs multiscale hypergraphs of varying group sizes to model groupwise correlations, captured through spectral hypergraph convolution based on random-walk probabilities. In parallel, a spatial-temporal transformer is employed to learn pedestrians’ pairwise latent interactions across spatial and temporal dimensions. These heterogeneous groupwise and pairwise features are subsequently fused and aligned via a multimodal transformer. Extensive experiments on public pedestrian motion datasets demonstrate that Hyper-STTN consistently outperforms state-of-the-art baselines and ablation models.


\end{abstract}




\section{Introduction}

\begin{figure}[!t]
\centering
\includegraphics[width=1\columnwidth]{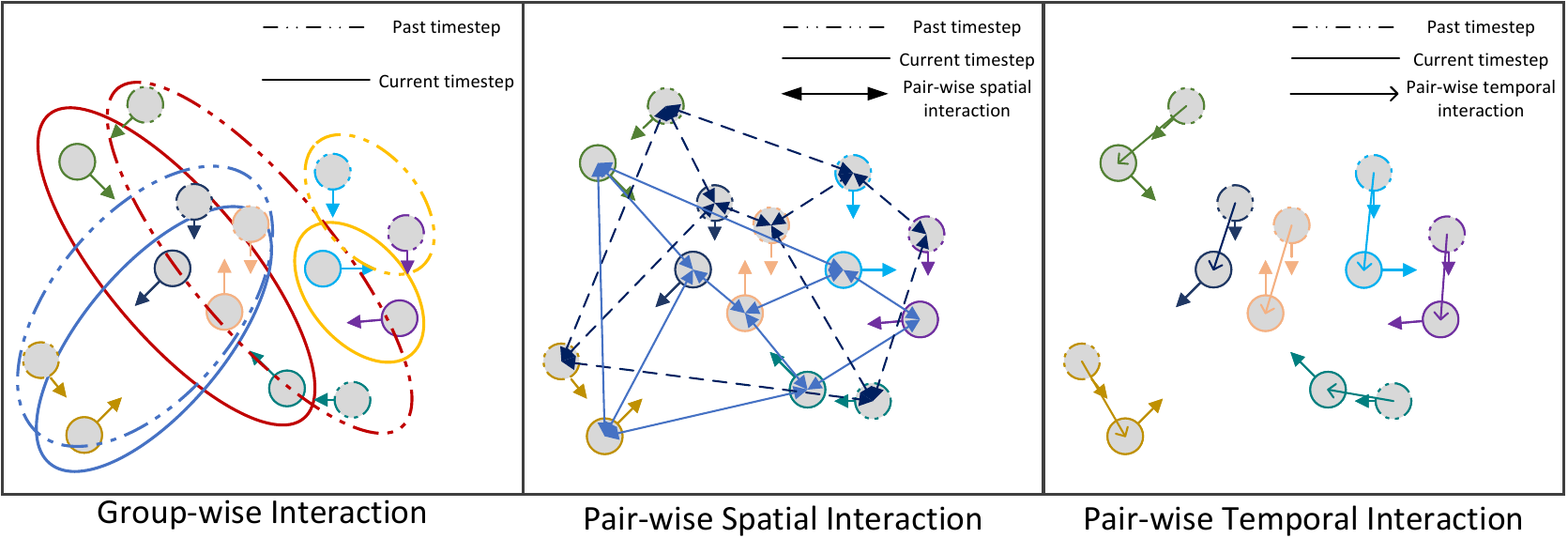}
\vspace{-20pt}
\caption{HHI feature illustration: groupwise HHI captures latent correlations among high-level perspectives on group behaviors, while pairwise spatial-temporal HHI represents individual influences.}
\vspace{-20pt}
\label{fig:F1}
\end{figure}

Human trajectory prediction is a pivotal research topic in computer vision and robotics, aiming to anticipate agents' future movements based on their past behaviors. The forecasting capability is critical for many real-world applications such as smart city systems and social robot navigation \cite{wang2023navistar, wang2023multi, wang2025hypergraph, wang2025human}. However, accurately forecasting human trajectories in social environments remains highly challenging due to the inherent stochasticity and unpredictability, and the diversity of crowd dynamics. These difficulties are compounded by the need to model intricate interactions among individuals and groups, involving both cooperative and competitive social behaviors, as illustrated in Fig.~\ref{fig:F1}.

Human trajectory forecasting is governed by three interrelated factors: individual intrinsic states, pairwise and groupwise social interactions, and instantaneous intentions. However, lots of existing approaches have struggled in modeling higher-order interactions and in reasoning across heterogeneous feature domains. Specifically, individual intrinsic states encode correlations between an agent’s current conditions and temporal patterns, implying future velocities or locations to be inferred from historical observations through sequence reasoning processes, as demonstrated in prior studies \cite{sutskever2014sequence,alahi2016social}. Moreover, contemporary forecasting frameworks are still unable to accurately infer subjective intentions from sparse or ambiguous observations, even when employing advanced machine learning or neural network architectures \cite{xu2022groupnet}. Thus, addressing these challenges necessitates enhanced representations of latent social dynamics, achieved through the joint integration of group-level embeddings and pairwise features across spatial-temporal dimensions.


Recently, recent research has increasingly focused on estimating complex social influences among humans to better capture the uncertainty of human movements. For instance, \cite{alahi2016social, gupta2018social, yu2020spatio, yuan2021agentformer} employ neural networks and attention mechanisms to model pairwise social interactions among agents across spatial-temporal dimensions. Nevertheless, the absence of an explicit graph structure constrains the expressiveness of the extracted features. Alternatively, some works \cite{yu2020spatio, salzmann2020trajectron++} incorporate spatial-temporal graphs, in which spatial edges encode interactions between agents and temporal edges represent individual temporal dependencies reflecting intrinsic states. More recently, \cite{2020Evolve, xu2022groupnet} extended this paradigm by constructing pedestrian group dynamics as a hypergraph to capture high-order group-level interactions for trajectory forecasting, where each hyperedge connects multiple vertices to represent collective group dynamics.


Despite state-of-the-art (SOTA) algorithms have achieved remarkable performance, inadequate modeling of human-human interaction (HHI) continues to constrain further advances on forecasting accuracy, particularly in highly dynamic environments. None of the above approaches fully model HHI across both group-level and pairwise feature dimensions \cite{alahi2016social}, nor do they effectively align heterogeneous multimodal dependencies \cite{yu2020spatio}. For instance, the lack of group-level interaction reasoning may obscure intergroup coordination or conflict \cite{salzmann2020trajectron++}. In team sports such as basketball, different defensive strategies exemplify this challenge: zone defenses typically involve one offensive player interacting with multiple defenders, whereas man-to-man schemes entail predominantly pairwise matchup. Furthermore, insufficient spatial-temporal feature inference can exacerbate ambiguities in HHI modeling \cite{xu2022groupnet}.

To address the aforementioned challenges, we propose Hyper-STTN, a hypergraph-based spatial-temporal transformer network explicitly designed to model both pairwise and groupwise social interactions across spatial-temporal dimensions, as illustrated in Fig.~\ref{fig:F1}. The central idea of Hyper-STTN is to provide an effective framework for inferring HHI and social dependencies by leveraging multiscale hypergraphs. The main contributions of this work are as follows: (1). We introduce Hyper-STTN, which constructs crowd dynamics using a set of multiscale hypergraphs to jointly capture groupwise and pairwise social interactions in both spatial and temporal domains; (2) Hyper-STTN incorporates a multimodal transformer module to align heterogeneous features and interactions, thereby mitigating interpretational ambiguities; (3). Extensive experiments on publicly available pedestrian trajectory datasets demonstrate that Hyper-STTN outperforms existing SOTA approaches.

\vspace{-5pt}
\section{Background}
\vspace{-5pt}
\subsection{Related Works}
Early efforts to human trajectory prediction relied on handcrafted models, such as the social force model \cite{yamaguchi2011you} and Gaussian processes \cite{wang2007gaussian}, which leverage fixed physical or mathematical rules to abstract environmental dynamics. However, the inherent inflexibility of these traditional methods leads to overlooking latent social interactions and producing inaccurate forecasts under complex conditions. Motivated by advances in machine learning, numerous deep learning–based algorithms have been developed to further express social interactions among pedestrians \cite{mao2023leapfrog, fu2025moflow, alahi2016social, gupta2018social}. However, previous learning-based methods either fail to adaptively incorporate the extracted features or cannot effectively leverage long-term dependencies. Inspired by the success of attention mechanisms and transformers in sequence learning and pairwise feature representation \cite{vaswani2017attention, yu2020spatio, vemula2018social}, recent work has explicitly modeled social interactions across both spatial and temporal dimensions through spatial-temporal graph representations of crowd movements. While pairwise interactions have been effectively addressed through the development of transformers, aforementioned approaches still neglect to capture groupwise latent influences on individual trajectory.

Despite prior studies \cite{2020Evolve, xu2022groupnet} have adapted hypergraph-based networks to learn groupwise features, they still struggles to disentangle spatial interactions along the temporal dimension. Consequently, aligning groupwise and pairwise social interactions with long-term dependencies remains a critical challenge in human trajectory prediction. In this work, we address this gap by not only integrating spatial and temporal social interactions from the mask attention mechanism into the hypergraph-based interaction representation, but also introducing the cross-modal attention mechanism \cite{tsai2019multimodal} to effectively exploit groupwise and pairwise heterogeneous multimodal dependencies.

\subsection{Social Interaction Reasoning}

Conventional sequence learning models, such as convolutional networks \cite{mohamed2020social, mohamed2022social} and recurrent networks \cite{alahi2016social, vemula2018social, salzmann2020trajectron++}, typically adopt hierarchical architectures that process information sequentially through network depth orderly. The lack of parallelization can potentially cause information to gradually vanish or explode across many layers of computation, particularly when capturing long-term dependencies. Conversely, transformer networks directly model element-to-element correlations in a fully parallel manner, leading to substantial improvements in sequence learning and pairwise interaction representation, as exemplified by \cite{vaswani2017attention}.

More recently, some existing works \cite{yu2020spatio, yuan2021agentformer, mao2023leapfrog, fu2025moflow} have employed transformer architectures to encode pairwise interactions across both spatial and temporal dimensions in trajectory prediction tasks. Given that agents' future trajectories are strongly correlated with their self-spatial and temporal features, those methods have achieved improvements over the SOTA performance via the better latent HHI feature representation. Nevertheless, although transformers can emphasize the importance of each pair of elements through attention mechanisms, they still tend to overlook groupwise features that capture higher-order crowd correlations.

Whereas standard graphs restrict edges to pairwise connections, hypergraphs provide a natural and expressive framework for modeling groupwise interactions, with hyperedges capable of simultaneously linking multiple vertices. Hypergraph models have been applied across diverse domains including biology, physics, recommender systems, and trajectory prediction \cite{grilli2017higher, ICMLHypergraph}. For instance, in ecological systems the correlations among predator-competitor groups are rarely linear. Especially, multiple predator-competitor interactions are difficult to approximate with pairwise edges alone. Accordingly, \cite{grilli2017higher} introduced higher-order interactions among biotic communities via hypergraphs to better capture and stabilize natural ecosystems. In a similar vein, \cite{xu2022groupnet} employed hypergraphs to capture groupwise social interactions for forecasting human trajectories. However, most existing approaches either rely on pre-defined topologies or construct hypergraphs solely from direct distances, without accounting for the underlying data distribution. In contrast, Hyper-STTN explicitly incorporates both pedestrian distribution effects and group scale to generate multiscale hypergraphs, while simultaneously embedding crowd dependencies across spatialtemporal dynamics within the hypergraph structure.


\section{Methodology}
\begin{figure*}[!t]
\centering
\includegraphics[width=0.97\linewidth]{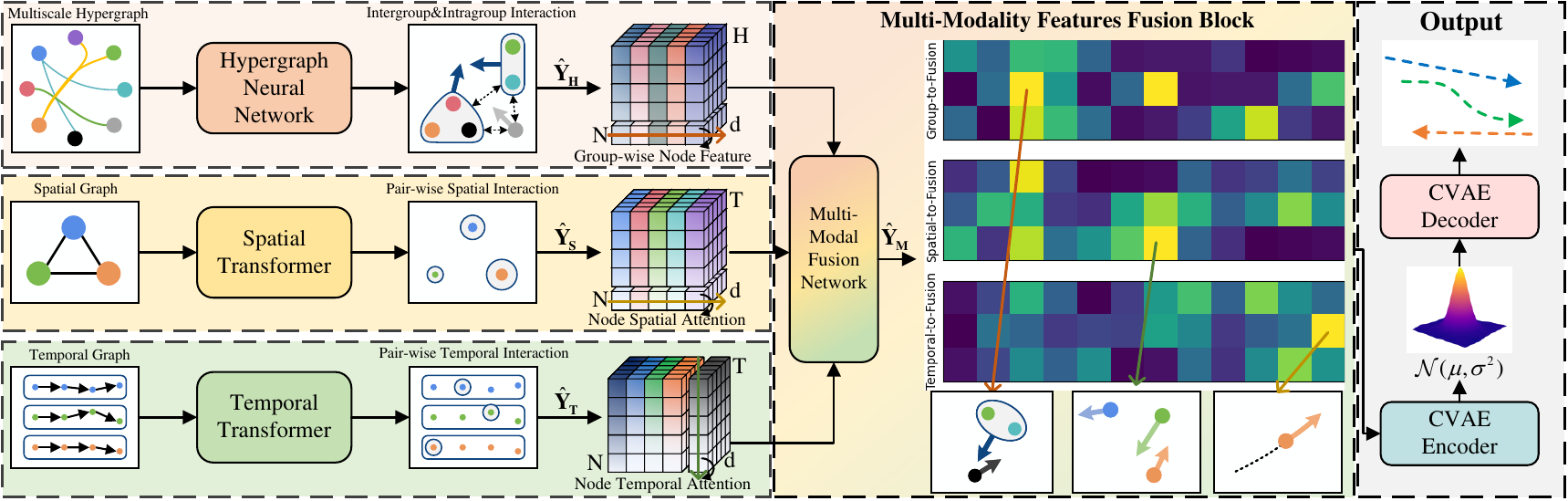}
\vspace{-10pt}
\caption{Hyper-STTN neural network framework: (a) Spatial Transformer leverages a multi-head attention layer and a graph convolution network along the time-dimension to represent spatial attention features and spatial relational features; (b) Temporal Transformer utilizes multi-head attention layers to capture each individual agent's long-term temporal attention dependencies; and (c) Multi-Modal Transformer fuses heterogeneous spatial and temporal features via a multi-head cross-modal transformer block and a self-transformer block to abstract the uncertainty of multimodality crowd movements.}
\vspace{-10pt}
\label{fig:F2}
\end{figure*}
\subsection{Preliminary}


We construct a set of multiscale hypergraphs $\mathcal{G} = (\mathcal{V,E,T}\mathbf{, W})$ for the representation of crowd movement dynamics. Where $\mathcal{V} = \{{v_{\rm 1}},{v_{\rm 2}},\cdots,{v_{\rm N}}\}$ denotes the set of vertices presenting the ${\rm N}$ agents in the scenario, and $\mathcal{E} = \{{e_{\rm 1}},{e_{\rm 2}},\cdots,{e_{\rm M}}\}$ is the set of ${\rm M}$ hyperedges among such vertices. Each hyperedge can link more than two vertices to form the groupwise interaction. The set $\mathcal{T} = \{{\tau_{\rm 1}},{\tau_{\rm 2}},\cdots,{\tau_{\rm H}}\}$ specifies total $\rm H$ multiscale hypergraphs. The weight matrix of each hyperedge is defined as $\mathbf{W} = diag({{w}_{\rm e_1}},{{w}_{\rm e_2}},\cdots,{{w}_{\rm e_M}}) \in \mathbb{R}^{M \times M}$. Moreover, the incident matrix $\mathbf{H} \in \mathbb{R}^{N \times M}$ of hypergraph $\mathcal{G}$ is defined as follows: $\mathbf{H}(v,e) = 1, ~\text{if}~ v \in e;~ \mathbf{H}(v,e) = 0, ~ \text{if} ~ v \notin e.$ The diagonal vertex matrix $\mathbf{M}_v \in \mathbb{R}^{N \times N}$ and diagonal edge matrix $\mathbf{M}_e \in \mathbb{R}^{M \times M}$ are composed by the degree of vertex $d(v)=\sum_{e\in \mathcal{E}}w_{(e)}\mathbf H(v,e)$ and the degree of edge $d(e)=\sum_{v\in \mathcal{V}}\mathbf H(v,e)$, separately.

Additionally, the adjacency matrix of hypergraph $\mathcal{G}$ is given by $\mathbf{A}\in \mathbb{R}^{N \times N}$, where $\mathbf{A}=\mathbf{H} \mathbf{W} \mathbf{H}^{\top} - \mathbf{M}_{\rm v}$. Notably, Hyper-STTN not only enables the modeling of high-order HHI features via multiscale hypergraphs, but also represents the pairwise dependencies with respect to the lowest-scale hypergraph $\tau_1$ that degenerates into a standard graph in the scale-1. Let $\mathbf{X} \in \mathbb{R}^{N \times T_{i} \times d}$ and $\mathbf{\hat{X}} \in \mathbb{R}^{N \times T_{o} \times d}$ denote, separately, the past and future trajectory data of crowd, where $T_i=8$ and $T_o=12$ are the input and output sequence lengths with the dimension $d = 2$. The position of $i-$th agent at time $t$ is expressed as $\mathbf{x}_{i}^{t} = (x,y)$. Eventually, Hyper-STTN describes the pedestrian trajectory distribution $\mathcal{P}$ as:
\begin{equation}
\{\mathbf{\hat{X}}_{1}, \cdots ,\mathbf{\hat{X}}_{N}\}= \mathcal{P}(\{\mathbf{X}_{1}, \cdots ,\mathbf{X}_{N}\};\mathcal{G})
\end{equation}
\noindent where $\{\mathbf{X}_{1}, \cdots ,\mathbf{X}_{N}\}$ and$\{\mathbf{\hat{X}}_{1}, \cdots ,\mathbf{\hat{X}}_{N}\}$ are the complete sets of input and output sequences for all agents in the scenario. For the $i-$th agent, the input sequence and corresponding output sequence are $\mathbf{X}_{i}=\{\mathbf{x}_{i}^{-{T}_{i}+1}, \cdots ,\mathbf{x}_{i}^{0}\}$, $\mathbf{\hat{X}}_{i} = \{\mathbf{{x}}_{i}^{1}, \cdots ,\mathbf{{x}}_{i}^{{T}_{o}}\}$.

\subsection{Hyper-STTN Architecture}
Hyper-STTN captures both groupwise and pairwise interactions to reason out HHI for trajectory prediction, as illustrated in Fig.~\ref{fig:F2}. Specifically, the spatial-temporal transformer network and the hypergraph convolution neural network are leveraged in parallel to encode pairwise HHI and groupwise HHI features, constructing a set of multiscale crowd hypergraphs. Subsequently, the heterogeneous interaction embeddings are fused via the cross-attention from the multimodal transformer. Eventually, forecasting trajectories are then decoded by a CVAE decoder \cite{sohn2015learning}.


\subsection{Spatial-Temporal Transformer}

\begin{figure}[!t]
\centering
\includegraphics[width=0.95\columnwidth]{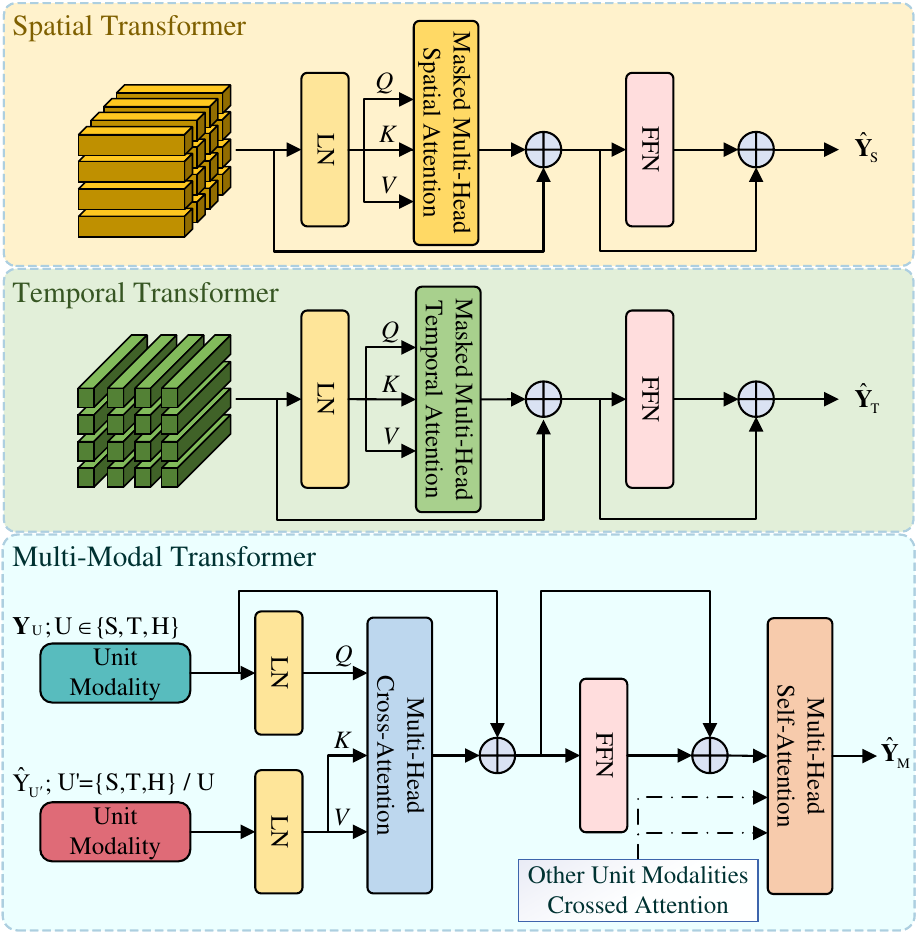}
\vspace{-10pt}
\caption{Hybrid Spatial-Temporal Transformer Framework: Pedestrians' motion intents and dependencies are abstracted as spatial and temporal attention maps by multi-head attention mechanism of spatial-temporal transformer. Additionally, a multi-head cross attention mechanism is employed to align heterogeneous groupwise and pairwise features.}
\vspace{-15pt}
\label{fig:F4}
\end{figure}

Drawing inspiration from \cite{chen2022bidirectional, yu2020spatio, wang2023navistar, mao2023leapfrog}, we designed a spatial-temporal transformer network to abstract the correlation of pairwise agents across spatial-temporal dimensions. Whereas spatial and temporal dependencies are represented by the relative importance in the attention maps (Fig.~\ref{fig:attention}). For the spatial transformer, the input sequence $\mathbf{X}$ is first processed by a positional encoding layer to incorporate sequence information \cite{vaswani2017attention}. The pre-processed data then passes sequentially through a layer normalization block (LN), a multihead mask attention layer, and a feed-forward network (FFN) (Fig.~\ref{fig:F4}), with residual connections employed to stabilize training. At the core of the spatial-temporal transformer is the attention mechanism, which the vanilla multihead attention mechanism is defined as follows:
\begin{equation}
\vspace{-5pt}
\begin{aligned}
&\operatorname{Atten}\left(\rm \mathbf Q_{\rm{}}^{i},\rm \mathbf K_{\rm{}}^{i},\rm \mathbf V_{\rm{}}^{i}\right)={\mathrm{softmax} (\frac{ \rm \mathbf Q_{\rm{}}^{i} \rm \mathbf ({\mathbf{K}_{\rm{}}^{i}})^{\top}}{\sqrt{ \rm{d}_{{h}}}}) \rm \mathbf V_{\rm{}}^{i}}\\
&\operatorname{Multi}\left(\mathbf Q_{\rm{}}^{\rm i},\mathbf K_{\rm{}}^{\rm i},\mathbf V_{\rm{}}^{\rm i}\right)=f_{\rm{fc}}( \rm{head}_1,\cdots, \rm{head}_{\rm{h}} )\\
&\rm{head}_{(\cdot)}=\operatorname{Atten}_{(\cdot)}\left(\rm \mathbf{Q}_{\rm{}}^{i}, \rm \mathbf{K}_{\rm{}}^{i}, \rm \mathbf{V}_{\rm{}}^{i}\right)
\end{aligned}
\end{equation}
where $f_{\rm{fc}}$ is a fully connected layer, $\rm{head}_{(\cdot)}$ denotes $\rm i$-th head in the multihead attention with $\rm{i} \in [1,\cdots,\rm{h}]$, $\rm \mathbf{Q}_{\rm{}}^{i}, \rm \mathbf{K}_{\rm{}}^{i}, \rm \mathbf{V}_{\rm{}}^{i}$ are $\rm i$-th query matrix, key matrix, and value matrix with the dimension $\rm d_{h}$. 
\begin{figure*}[!t]
\centering
\includegraphics[width=0.96\linewidth]{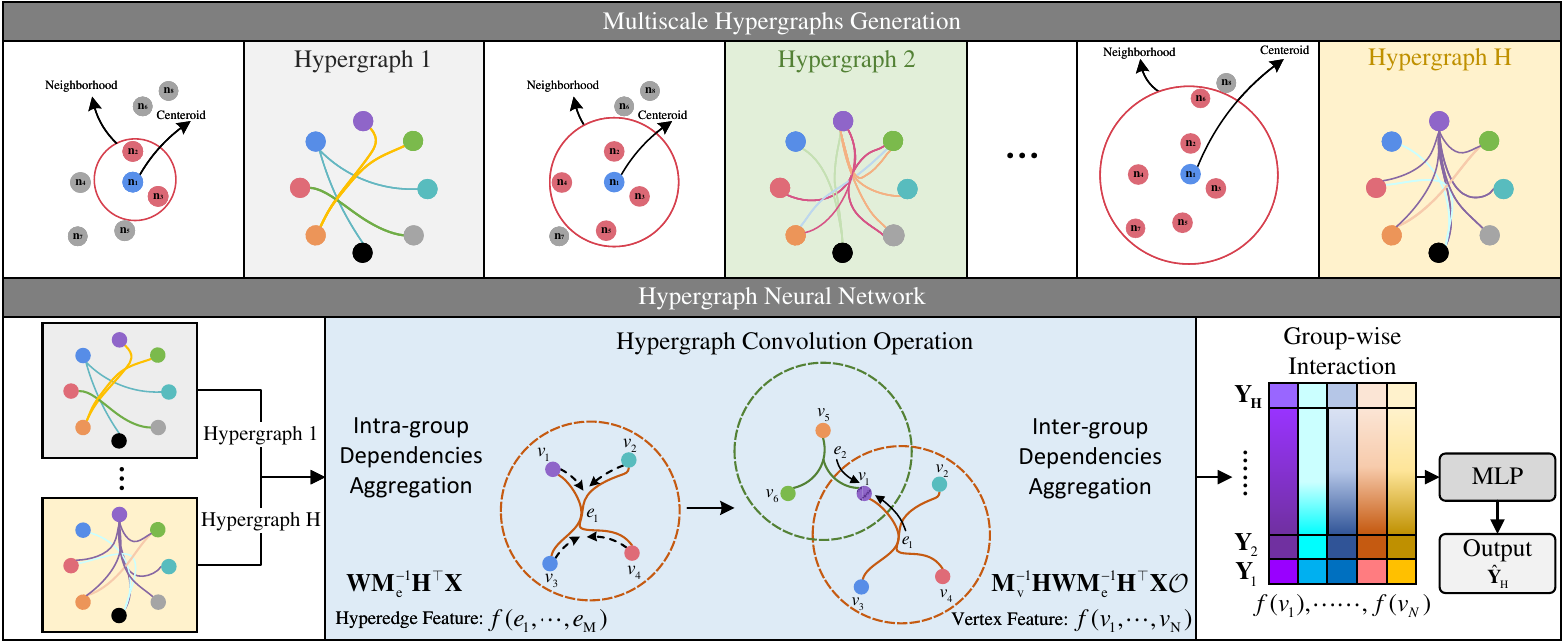}
\vspace{-10pt}
\caption{Groupwise HHI Representation: i) We construct groupwise HHI with a set of multiscale hypergraphs, where each agent is queried in the feature space with varying 'k' in KNN to link multiscale hyperedges. ii) After constructing HHI hypergraphs, groupwise dependencies are captured by point-to-edge and edge-to-point phases with hypergraph spectral convolution operations.}
\vspace{-10pt}
\label{fig:F3}
\end{figure*}
Additionally, due to limitations in data collection, certain timesteps of individual trajectories may be unrecorded in the temporal transformer, while some pedestrians may appear intermittently in the spatial transformer. To address this variability, a masked attention mechanism \cite{cheng2022masked} is employed, enabling the spatial–temporal transformer to robustly handle length-varying sequence data.
\begin{equation}
\begin{aligned}
&\operatorname{MAtten}\left(\rm \mathbf Q_{\rm{}}^{i},\rm \mathbf{K}_{\rm{}}^{i},\rm \mathbf V_{\rm{}}^{i}\right)={\mathrm{softmax} (\mathcal{M}^{\rm i} + \frac{\rm \mathbf Q_{\rm{}}^{i} \rm \mathbf ({K_{\rm{}}^{i}})^{\top}}{\sqrt{ \rm{d}_{{h}}}})  \rm \mathbf V_{\rm{}}^{i}}\\
\end{aligned}
\end{equation}
where attention mask matrix $\mathcal{M}$ is defined to handle the issue of varying length data and to encode relative distance feature $\omega(n,t)=f_{fc}(dis(\text{agent-pair}))$ into the attention as follows:
\begin{equation}
\mathcal{M}(\rm n,t) =\left\{
	\begin{aligned}
	-\infty \quad\quad &\mathbf{X}(\mathrm{n,t})={none}\\
 	\omega(n,t) \quad\quad &otherwise\\
	\end{aligned}
	\right.
\end{equation}
The framework of temporal transformer and spatial transformer are illustrated as Fig.~\ref{fig:F4}. The spatial transformer captures pairwise spatial interactions $\mathbf{\hat{Y}}_{\rm{S}} \in \mathbb{R}^{N \times T_{i} \times d}$ among all agents at each timestep, while the temporal transformer summaries individual temporal interactions $\mathbf{\hat{Y}}_{\rm{T}} \in \mathbb{R}^{N \times T_{i} \times d}$ for each agent.
\begin{equation}
\begin{aligned}
&\mathbf{\hat{Y}}_{\rm{S}};\mathbf{\hat{Y}}_{\rm{T}} =  \rm{Trans}_{\rm{spatial}}(\mathbf X);\rm{Trans}_{\rm{temporal}}(\mathbf X)
\end{aligned}
\vspace{-5pt}
\end{equation}

\subsection{Hypergraph Neural Network}
\vspace{-5pt}
The crowd groups of Hyper-STTN are constructed with respect to neighborhoods' feature dimensional Mahalonobis distance \cite{wang2015survey}, considering both interactive correlation distribution and physical spatial distance. In particular, the pedestrians' motion similarity is captured by the covariance matrix of Mahalonobis distance to cover the situations that a long-distance person presents a high potentiality to join the group. We construct crowd hypergraphs by vertex classifications of social group which are formulated as a spectral hypergraph $k$-way partitioning problem \cite{zhou2006learning, HGNN+}. We define a group at time $t$ as a set of agents whose latent motion embeddings are mutually similar under a Mahalanobis metric computed over short-horizon kinematics (position, velocity, heading) and context crowd distribution features; proximity alone is not sufficient. Hyperedges connect an agent to its top-$k$ neighbors under this metric; varying $k_i$ yields multiscale groups with respect to $\tau_i$. The normalized one cut spectral hypergraph partitioning task is defined as follows:



\vspace{-5pt}
\begin{equation}
\begin{aligned}
&\mathop{\arg\min}\limits_{f} \frac{1}{2} \sum_{e \in \mathcal{E}} \sum_{v_i, v_j \in \mathcal{V}} \frac{w(e)}{d(e)} [\frac{f(v_i)}{\sqrt{d(v_i)}} - \frac{f(v_j)}{\sqrt{d(v_j)}}]^{2} \\
&= \mathop{\arg\min}\limits_{f} f^{\top} \Delta f
\end{aligned}
\end{equation} 
\vspace{-2pt}
where $f(\cdot)$ is a classification function, and $\Delta$ is the positive semi-definite hypergraph Laplacian.

Assuming the $k$-way hypergraph partition by a set of vertices subsets $\{ \mathcal{V}_1,\cdots,\mathcal{V}_p\}$ from $\mathbf{F} = [f_1 \cdots f_p]$. Subsequently, the spectral hypergraph $k$-way partitioning problem as a combinatorial optimization problem can be relaxed with respect to minimizing $\beta(\mathcal{V}_1,\cdots,\mathcal{V}_p)$ as follows:
\begin{equation}
\mathop{\arg\min}\limits_{\beta} \{ \beta(\mathcal{V}_1,\cdots,\mathcal{V}_p) \} = \{ \sum_{i=1}^{p} f_i^{\top} \Delta f_i \} = \{ trace (\mathbf{F}^{\top}\Delta \mathbf{F}) \}
\end{equation} 

To address above k-way hypergraph partition task, the K-nearest neighbor (KNN) \cite{cunningham2021k} method is leveraged for multiscale hypergraphs generation. Hyper-STTN iteratively applies KNN to each vertex to identify its interactive neighborhoods based on the Mahalanobis feature distance. Particularly, the hypergraph at scale $\tau_1 =2$ degenerates to a standard graph, representing only pairwise interaction. In details, the low-level trajectory embeddings of each agent $\{\mathbf{q}_{(\rm{X}_{\emph{1}})}, \cdots, \mathbf{q}_{(\rm{X}_{\emph{N}})}\}$, where $\mathbf{q}({x}_{i}) \in {\mathbb{R}}^{d}$ are obtained via a fully connected layer from the input sequences $\{\mathbf{X}_{\emph{1}}, \cdots ,\mathbf{X}_{\emph{N}}\}$. The similarity matrix $\mathcal{S}$ is then filled by feature dimensional agent pairs' Mahalanobis distance \cite{wang2015survey}, which incorporate interaction attributes and motion distributions to mitigate Euclidean distance based distortions arising from feature correlations and heterogeneous distributions. The Mahalanobis distance of $(i,j)$-th vertex pair $Dis(i,j)$ is defined as follows:
\begin{equation}
Dis(i,j) = \sqrt{[\mathbf{q}_{(\rm{X}_{\emph{i}})} - \mathbf{q}_{(\rm{X}_{\emph{j}})}]^{\top} {\sum}^{-1}[\mathbf{q}_{(\rm{X}_{\emph{i}})} - \mathbf{q}_{(\rm{X}_{\emph{j}})}]}
\end{equation}
\noindent where ${\sum}^{-1}$ is the covariance matrix of sample distribution. 

The $(i,j)$ element of similarity matrix $\mathcal{S} \in {\mathbb{R}}^{N \times N}$ is defined as follows:
\begin{equation}
\mathcal{S}(i,j) = \exp [-\frac{Dis(i,j)^2}{\varrho^2}]
\end{equation}
\noindent where $\varrho$ is the mean of all vertex pairs' feature distances.

Inspired by \cite{feng2019hypergraph, HGNN+}, we initialize multiscale hypergraphs by performing KNN searches for each vertex and its $k$ nearest neighbors with respect to the similarity matrix $\mathcal{S}$, as shown in Fig.~\ref{fig:F3}. In this representation, each vertex corresponds to a single agent, while hyperedges link multiple agents for the representation of groupwise interaction. The $k$-nearest neighbor selection is kept fixed per training epoch from the current embeddings (stop-gradient on neighbor indices). We refresh neighbor sets every $E$ epochs to account for updated embedding structures during training.

After constructing multiscale crowd hypergraphs, a hypergraph convolution neural network \cite{HGNN+} is employed to estimate the groupwise interactions. The random walk probability \cite{zhou2006learning} of hypergraph aggregates the weighted dependencies of all sub-vertices into hyperedges features, which are then combined into vertex features, as shown in Fig.~\ref{fig:F3}. Let a vertex $v_i$ be stochastically selected with one of its hyperedges $e_o$, and the probability of hypergraph random walk $\mathcal{O}_{(v_i,v_j)}$ from $v_i$ to $v_j$ on hyperedge $e_o$ can be defined as follows:
\begin{equation}
\mathcal{O}_{(v_i,v_j)}={\sum}_{e \in \mathcal{E}}\mathbf{W}_{e_o} \frac{\mathbf{H}(v_i,e_o) \mathbf{H}(v_j,e_o)}{\mathbf{M}_{v}(i,i) \mathbf{M}_{e}(o,o)}
\end{equation}

The matrix normalized form of hypergraph random walk can be expressed as follows:
\begin{equation}
\mathcal{O}= \mathbf{M}_{v}^{-\frac{1}{2}} \mathbf{H} \mathbf{W} \mathbf{M}_{e}^{-1} \mathbf{H}^{\top} \mathbf{M}_{v}^{-\frac{1}{2}}
\end{equation}

Based on the definition of random walk probability, the hypergraph interaction features, representing groupwise interaction information are calculated by a hypergraph spectral convolution operation grounded, which aggregates the global dependencies between vertices and hyperedges. Following \cite{kipf2017semisupervised, HGNN+}, the spectral hypergraph convolution of an input signal $\mathbf{x} \in \{\mathbf{x}_1,\cdots,\mathbf{x}_N\}$ with a filter $\mathbf{g}$ is defined as follows:
\begin{equation}
\mathbf{g} \otimes \mathbf{x} := {{\sum}^{U}_{u=0}} \theta_{u} \mathbf{T}_{u}(\widetilde{\Delta}) \mathbf{x}
\end{equation}
where $\Delta = \mathbf{I} - \mathcal{O}$ is the regularized hypergraph Laplacian matrix. And $\mathbf{T}_{u}(\widetilde{\Delta})$ is the Chebyshev polynomial of order $u$ with
scaled Laplacian $\widetilde{\Delta}= 2 {\Delta} / {\lambda_{\max }}-\mathbf{I}$, where ${\lambda_{\max }}$ is the largest eigenvalue of $\Delta$. The parameter $\theta$ is a weighted parameter, and ${U}$ is the kernel size of the graph convolution.

Decomposing the hypergraph Laplacian matrix and approximating it by the first-order Chebyshev polynomial as a hypergraph convolution operation with single scale groupwise interaction dependence as output $\mathbf{Y} \in \mathbb{R}^{N \times d}$.
\begin{equation}
\mathbf{Y} = \mathbf{M}_{v}^{-1} \mathbf{H} \mathbf{W} \mathbf{M}_{e}^{-1} \mathbf{H}^{\top} \mathbf{X} \mathcal{O}
\end{equation}

Finally, we aggregate all the multiscale hypergraphs together to present crowded miscellaneous groupwise interactions as follows:
\begin{equation}
\mathbf{\hat{Y}_{\rm{H}}} = f_{MLP}[Concat^{H}_{h=1}(\mathbf{Y})]
\end{equation}
where $f_{MLP}$ is a MLP neural network, and $\mathbf{\hat{Y}_{\rm{H}}} \in \mathbb{R}^{N \times H \times d}$ is the groupwise interaction feature.

\subsection{Multi-Modal Fusion Network and CVAE Decoder}

As shown in Fig.~\ref{fig:F4}, the multi-modal transformer \cite{tsai2019multimodal} is developed to fuse heterogeneous spatial-temporal features, using a cross-attention layer and self-attention layer. Wherein the multihead cross-attention mechanism $\rm CMAtten(\cdot)$ captures cross-modality features as follows:
\begin{equation}
\begin{aligned}
\rm{CMAtten}(\hat{Y}_U)
& = \rm{Multi}(\mathbf Q^{\rm{head_j}}_{\rm{U}}, \mathbf K^{\rm{head_j}}_{\rm{U'}}, \mathbf V^{\rm{head_j}}_{\rm{U'}})\\
\end{aligned}
\end{equation}
where $\rm U \in \{\rm{S}, \rm{T}, \rm{H} \}$, and $\mathbf Y^{\rm{head_j}}_{\rm{U}}$ present the $\rm j$-th head crossmodal attention of arbitrary unit modal with $\rm j \in \{1,\cdots,h\}$. 


Subsequently, the groupwise HHI features $\mathbf{\hat{Y}}_{\rm{HS}}, \mathbf{\hat{Y}}_{\rm{HT}}$ and pairwise HHI dependencies $\mathbf{\hat{Y}}_{\rm{ST}}$ are aligned by a multi-head self-attention network as final crowd dynamics representation $\mathbf{\hat{Y}}_{\rm{M}} =  \rm{Trans}_{\rm{multimodal}}(\mathbf{\hat{Y}_{\rm{H}}}, \mathbf{\hat{Y}_{\rm{S}}}, \mathbf{\hat{Y}_{\rm{T}}})$, where neural network $\rm{Trans}_{\rm{multimodal}}$ refers to \cite{tsai2019multimodal}.

Eventually, to estimate the stochasticity of human movements, a conditional variational auto encoder (CVAE)-based decoder \cite{sohn2015learning} is employed to approximate maximum likelihood in potential distributions of motion uncertainty. The environmental dynamics feature $\mathbf{\hat{Y}}_{\rm M}$ and observed data $\mathbf{X}$ are encoded to present Gaussian distribution ${ \mathcal{N}(\mu,\sigma^2)}$. And the latent variable $z$ is sampled by above Gaussian distribution $z \sim {\mathcal{N}(0,\sigma_{\rm T_i}^2\mathbf{I})}$ in the testing process. Lastly, the maximum likelihood forecasting result is calculated by the decoder based on the concatenation of latent variable and observation embedding in the CVAE block as follows:
\begin{equation}
\begin{aligned}
 z &\sim \{ \mathcal{N}(\mu,\sigma^2) =Encoder(\mathbf X,\mathbf{\hat{Y}}_{\rm M}) \}\\
 \mathbf{\hat{X}} &\sim Decoder(\mathbf{\hat{X}} \; | \; z,Encoder(\mathbf X))
\end{aligned}
\end{equation}
where $\mu,\sigma$ are the mean and variance of the approximate distribution, and $Encoder(\cdot)$ is Hyper-STTN backbone framework, and $Decoder(\cdot)$ is a spectral temporal graph network from \cite{cao2020spectral}.

\section{Experiments And Results}

\subsection{Experiments}
\subsubsection{Dataset}
We conduct experiments on several widely used public pedestrian trajectory datasets collected from real-world scenarios. The training and validation data comprise the ETH-UCY and NBA \cite{xu2022groupnet} datasets, which record pedestrians’ temporal location information in world coordinates. Consistent with prior works, our experiments adopt a leave-one-out cross-validation pre-process method, reserving only the last subsection of each scene for testing.




\subsubsection{Evaluation Metrics}
In our experiments, we evaluate prediction accuracy using the average displacement error (ADE) and final displacement error (FDE) metrics \cite{alahi2016social}. The ADE$_{20}$/FDE$_{20}$ quantify performance with the Euclidean distance between predicted and ground-truth trajectories, either averaged over all timesteps or at the final timestep, which evaluate performance on the top 20 sampled predictions.
\vspace{-5pt}
\subsection{Comparison Configuration}
\subsubsection{Baselines}
We compared our model with several existing SOTA algorithms. The algorithms included in the comparison are Social-Attention \cite{vemula2018social}, Social-GAN \cite{gupta2018social}, Social-STGCNN \cite{mohamed2020social}, Trajectron++ \cite{salzmann2020trajectron++}, STAR \cite{yu2020spatio}, PECNet \cite{mangalam2020not}, Social-Implicit \cite{mohamed2022social}, GroupNet \cite{xu2022groupnet}, EqMotion \cite{xu2023eqmotion}, LED \cite{mao2023leapfrog}, and IMLE \cite{fu2025moflow}, as shown in Table~\ref{table: Stochastic} and Table~\ref{table: NBA}.


\subsubsection{Ablation Models}
We design two ablation models to evaluate the contributions of distinct components for overall performance. The ablation model STTN, infers pedestrian interactions using only the hybrid spatial-temporal transformer, excluding the hypergraph network block from Hyper-STTN. The second mode, HGNN, estimates crowd motion dependencies solely through hypergraph convolution operations without transformers. In both ablation models, the CVAE structure is retained as the decoder. All training configurations, including datasets and initialization parameters, are kept identical to those used for Hyper-STTN.

\begin{table*}[h]
\caption{The crowded trajectory forecasting best-of-20 stochastic sampled results of minADE$_{20}$ / minFDE$_{20}$ on ETH-UCY dataset.}
\vspace{-6pt}
\resizebox{\textwidth}{!}{
\begin{tabular}{@{}c|@{ }c@{ }c@{ }c@{ }c@{ }c@{ }c@{ }c@{ }c@{ }c|@{ }c@{ }c|@{ }c}
	\hline\hline 
    \thead{\textbf{Stochastic}\\ADE$_{20}$ / FDE$_{20}$}  & \thead{Social-\\Attention \cite{vemula2018social}\\ICRA18} & \thead{Social-\\GAN \cite{gupta2018social}\\CVPR18} & \thead{Trajectron\\++ \cite{salzmann2020trajectron++}\\ECCV20} & \thead{STAR \\ \cite{yu2020spatio}\\ ECCV20} & \thead{Social-\\Implicit \cite{mohamed2022social}\\ECCV22} & \thead{GroupNet \\ \cite{xu2022groupnet} \\ CVPR22} & \thead{EqMotion\\ \cite{xu2023eqmotion}\\CVPR23} & \thead{LED \\ \cite{mao2023leapfrog} \\ CVPR23} & \thead{IMLE\\ \cite{fu2025moflow}\\CVPR25} & \thead{STTN\\(ablation)} & \thead{HGNN\\(ablation)}  & \thead{Hyper-STTN\\(ours)} \\
    \hline
    ETH   & 1.39/2.39 & 0.87/1.62 & 0.61/1.02  & {0.36}/0.65 & 0.61/1.08  & {0.46/0.73}  & 0.40/{0.61} & \bf{0.39/0.58}  & 0.40/\textbf{0.58}& 0.52/0.75 & 0.42/0.73 & \bf{0.35/0.57} \\
    
    HOTEL & 2.51/2.91 & 0.67/1.37 & 0.19/0.28  &  0.17/0.36 & 0.33/0.63  & 0.15/0.25  & {0.12/0.18} &  \bf{0.11/0.17}  &  0.12/0.18 & 0.21/0.25 & 0.21/0.32 & 0.14/0.18 \\
    
    UNIV  &  1.25/2.54 & 0.76/1.52 &  0.30/0.54  &  0.31/0.62 & 0.52/1.11 & 0.26/0.49 & \textbf{0.23}/{0.43} &  0.26/0.44 & \textbf{0.23/0.39} & {0.25/0.43} & 0.30/0.53 & \textbf{0.23}/{0.41} \\
    
    ZARA1 & 1.01/2.17 & 0.35/0.68 & 0.24/0.42  &  0.26/0.55 & 0.32/0.66 & 0.21/0.39 & {0.18}/0.32 &  0.18/\textbf{0.26} & \textbf{0.16/0.26} & 0.31/0.39 & 0.28/0.53 & {0.18}/{0.28} \\
    
    ZARA2 &  0.88/1.75 &  0.42/0.84 & 0.18/0.32  & 0.22/0.46 & 0.43/0.85 & 0.17/0.33 & \textbf{0.13}/0.23 &  \bf{0.13/0.22} & \bf{0.13/0.22} & 0.25/0.33 & 0.23/0.28 & 0.15/\bf{0.22} \\

    \textcolor{blue}{\textbf{AVG}}   & 1.41/2.35 & 0.61/1.21 & 0.30/0.51  & 0.26/0.53 & 0.33/0.67  & 0.25/0.44 & \textbf{0.21}/{0.35} & \textbf{0.21/0.33} & \textbf{0.21/0.33} & {0.31/0.43} & 0.29/0.48  & \textbf{0.21/0.33} \\
    
    \hline\hline
    \end{tabular}}
\label{table: Stochastic}
\vspace{-3pt}
\end{table*}


\begin{table*}[h]
\caption{The crowded trajectory forecasting best-of-20 stochastic sampled results of minADE$_{20}$ / minFDE$_{20}$ on NBA dataset.}
\vspace{-6pt}
\resizebox{\textwidth}{!}{
\begin{tabular}{@{}c|@{ }c@{ }c@{ }c@{ }c@{ }c|@{ }c@{ }c|@{ }c}
	\hline\hline
    \thead{\textbf{Stochastic}\\ADE$_{20}$ / FDE$_{20}$}  & \thead{Social-GAN \cite{gupta2018social}\\CVPR18} & \thead{Social-STGCNN \cite{mohamed2020social}\\CVPR20} & \thead{Trajectron++ \cite{salzmann2020trajectron++}\\ECCV20} & \thead{PECNet \cite{mangalam2020not}\\ ECCV20}  & \thead{GroupNet \cite{xu2022groupnet}\\CVPR22}    & \thead{STTN\\(ablation)} & \thead{HGNN\\(ablation)}  & \thead{Hyper-STTN\\(ours)} \\
    \hline
    
    NBA-1s   & 0.41/0.62 &  0.34/0.48 & \textbf{0.30/0.38}   & 0.35/0.58 & 0.34/0.48 & 0.39/0.58  & 0.37/0.59  & \textbf{0.30/0.37}  \\
    
    NBA-2s & 0.81/1.32 & 0.71/0.94 & \textbf{0.59/0.82}   & 0.68/1.23 & 0.62/0.95 &  0.72/1.05  & 0.64/0.95  & \textbf{0.58/0.81}  \\
    
    NBA-3s  &  1.19/1.94 & 1.09/1.77 &  \textbf{0.85/1.24}   & 1.01/1.76 & 0.87/1.31 & 1.07/1.51 & 0.98/1.44  & \bf{0.84}/\textbf{1.21}  \\
    
    NBA-4s & 1.59/2.41 &  1.53/2.26 & 1.15/\textbf{1.57}   & 1.31/1.79 & \textbf{1.13}/1.69 &  {1.24}/1.99 & {1.18}/1.89  & \textbf{1.01/1.52}  \\

\hline


\hline \hline
\end{tabular}}
\label{table: NBA}
\vspace{-5pt}
\end{table*}

\subsubsection{Training Details}
We use two NVIDIA RTX-4090 GPUs to train the model via the Adam optimizer. To improve robustness, Gaussian noise is injected during training. 


In our training procedure, we utilize the distance loss function $\mathcal{L}_{dis}$, angle loss function $\mathcal{L}_{ang}$, and encoder loss function $\mathcal{L}_{enc}$ to update the network as follows:
\begin{equation}
\begin{aligned}
&\mathcal{L}_{dis} = \kappa_1 \Vert {\mathbf{\hat{X}}^{}}-{\mathbb{{\hat{X}}}^{}} \Vert_{2} + \kappa_2 {f_{\mathrm{KL}}}(\mathcal{N}(\mu,\sigma^2)||\mathcal{N}(0,\sigma^2_{\rm T_i} \mathbf{I})) \\
&\mathcal{L}_{ang} = \kappa_3 \sum_{t \in T} \sum_{j \in T,j>t} \Vert \angle(\hat{\mathbf{X}}_t,\hat{\mathbf{X}}_j) - \angle({\mathbf{X}}_t,{\mathbf{X}}_j) \Vert_2 \\
&\mathcal{L}_{enc} = \kappa_4 \Vert {Encoder(\mathbf{X})}-{\mathbf{X}} \Vert_{2} \\
\end{aligned}
\end{equation}
where $\kappa$ is the weight of each loss term, and the Kullback-Leibler (KL) divergence term ${f_{\mathrm{KL}}(\cdot||\cdot)}$ is used to update the encoder and decoder in the CVAE block, and ${\mathbb{{\hat{X}}}^{}}$ denotes the groundturth of dataset. The $\angle(\cdot)$ function computes the angle between two vectors representing points.

Finally, the overall loss function for Hyper-STTN $\mathcal{L}$ is constructed as the sum of the above equations to minimize the total loss as follows: $\mathcal{L} =  \mathcal{L}_{dis} +  \mathcal{L}_{ang} +  \mathcal{L}_{enc}.$

\subsection{Quantitative Results}

\vspace{-5pt}



We conducted the trajectory prediction task with respect to stochastic conditions, performing Hyper-STTN's SOTA quantitative results in ETH-UCY and NBA datasets, as shown in Table~\ref{table: Stochastic} and Table~\ref{table: NBA}. Notably, due to the datasets’ size and the rapid advancement of expressive models, the performance on the ETH-UCY datasets has plateaued in recent years that is approaching the inherent errors of machine learning approaches on such datasets. However, Hyper-STTN still exhibits comparable performance against existing advanced models both on ADE$_{20}$ and FDE$_{20}$ metrics. For instance, in the ETH-UCY dataset, Hyper-STTN not only enhances the $12.5\%$ ADE$_{20}$ and $1.7\%$ FDE$_{20}$ from EqMotion's \cite{xu2023eqmotion} 0.40/0.58 to 0.35/0.57 on ETH dataset, but also improves the average $14.7\%/ 6.5\%/ 3.4\%/ 10.6\%$ ADE$_{20}$ and $22.9\% / 14.7\% / 7.6\% / 10.1\%$ FDE$_{20}$ performance from Groupnet's \cite{xu2022groupnet} on NBA dataset. Highlighting the effectiveness of Hyper-STTN in crowd latent interaction inference and trajectory forecasting.


Deep learning–based trajectory prediction frameworks have catalyzed a shift toward modeling latent social interactions across spatial and temporal dimensions for pedestrian motion forecasting. Methods built upon recurrent architectures \cite{alahi2016social}, attention mechanisms \cite{vemula2018social}, and, more recently, transformer models \cite{yu2020spatio, yuan2021agentformer} have demonstrated progressively stronger predictive performance by enabling richer high-level feature representations. Following this trend, Hyper-STTN adopts transformer backbones to explicitly model fundamental social interactions and temporal dependencies in dense crowds. Its ablation variant, STTN, which is purely transformer-driven (akin to STAR\cite{yu2020spatio, yuan2021agentformer}), isolates the contribution of these components. Experimental results indicate that the observed performance gains arise from the introduction of masked-attention and cross-modal attention modules. These modules mitigate agent-flicker effects relative to vanilla self-attention under fixed, predefined transformer data scales, while simultaneously improving the adaptability of spatial–temporal fusion.

Despite the remarkable effectiveness of transformers in abstracting pairwise interactions, their limited consideration of group-level dynamics constrains further performance improvements, particularly in dense scenes where individual motions are strongly influenced by the movements of surrounding groups. For example, GroupNet \cite{xu2022groupnet} achieves superior performance (0.26/0.49) compared to STAR \cite{yu2020spatio} (0.31/0.62) on the UNIV dataset, which contains more densely crowded scenarios. To address this gap, we propose to integrate hypergraph construction for inferring groupwise interactions with transformer architectures for pairwise feature modeling, thereby capturing richer strategies for crowded movement prediction. In addition, the ablation model HGNN exhibits performance comparable to GroupNet across several datasets, underscoring the effectiveness of the hypergraph component.

\vspace{-1pt}
\begin{figure}[!t]
\centering
\vspace{-8pt}
\hspace{-6pt}
\subfloat[Instance-1]{\label{fig:a} \includegraphics[width=4CM]{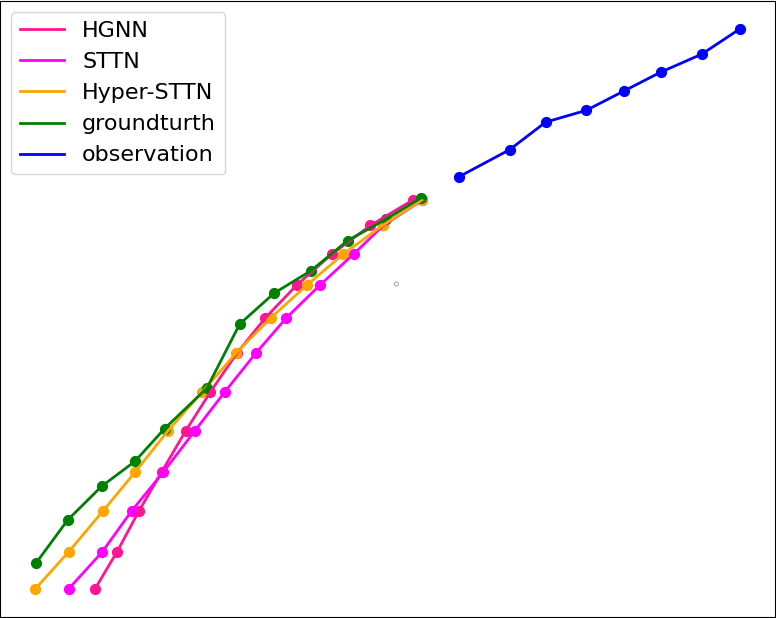}}~~
\subfloat[Instance-2]{\label{fig:b}\includegraphics[width=4CM]{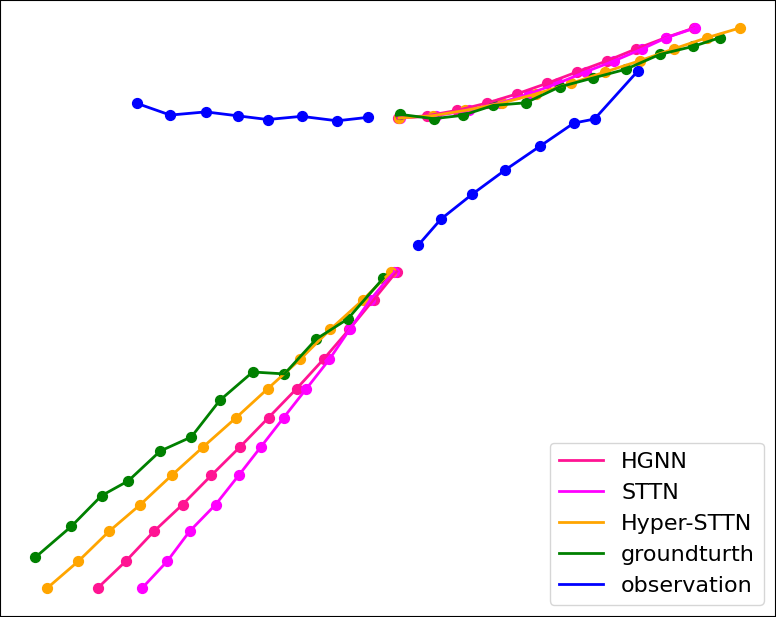}}
\vspace{-5pt}
\subfloat[Instance-3]{\label{fig:c}\includegraphics[width=4CM]{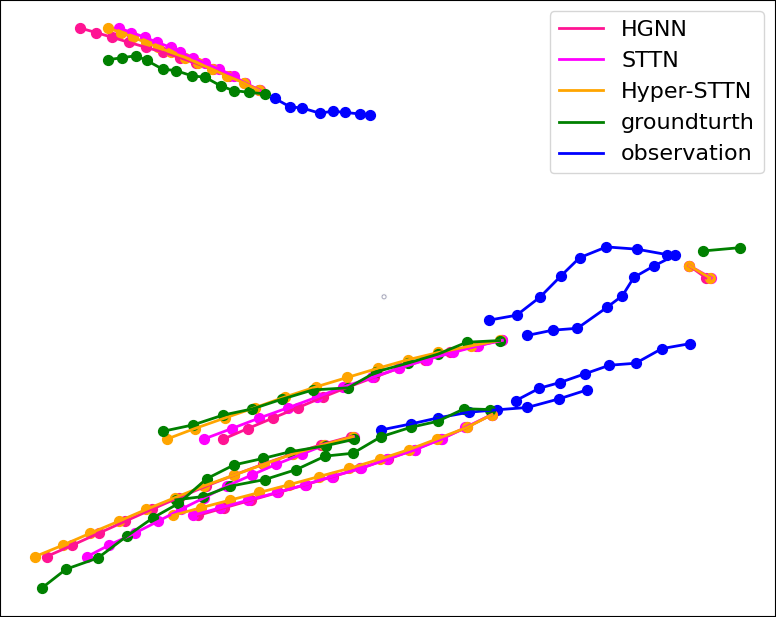}}~~
\subfloat[Instance-4]{\label{fig:d}\includegraphics[width=4CM]{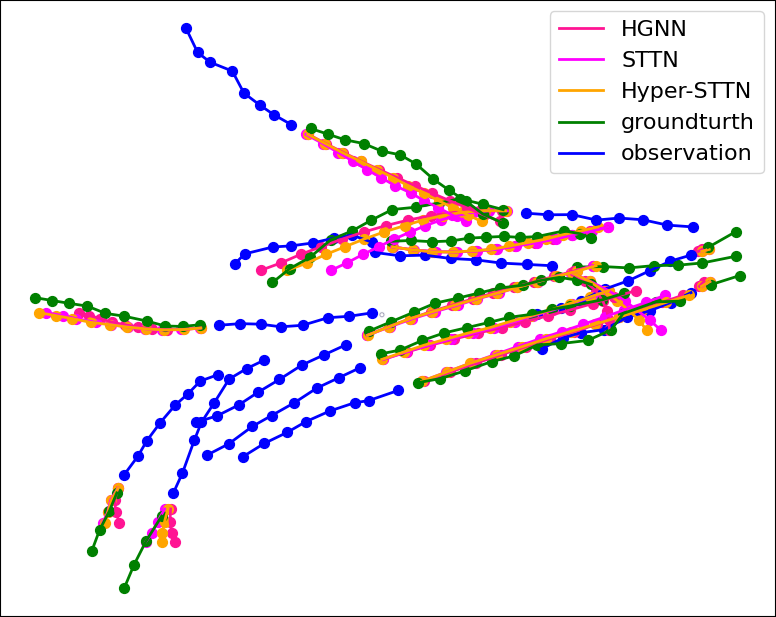}}
\vspace{-0pt}
\caption{Comparison of Trajectories Visualizations: The trajectories visualized for Hyper-STTN and other algorithms tested on the same test case.}
\vspace{-10pt}
\label{fig:traj}
\end{figure}

\vspace{-4pt}
\subsection{Discussion}
\begin{figure}[!t]
\centering
\vspace{-2pt}
\hspace{-6pt}
\subfloat[Spatial Attention Map]{\label{fig:sa} \includegraphics[width=4CM]{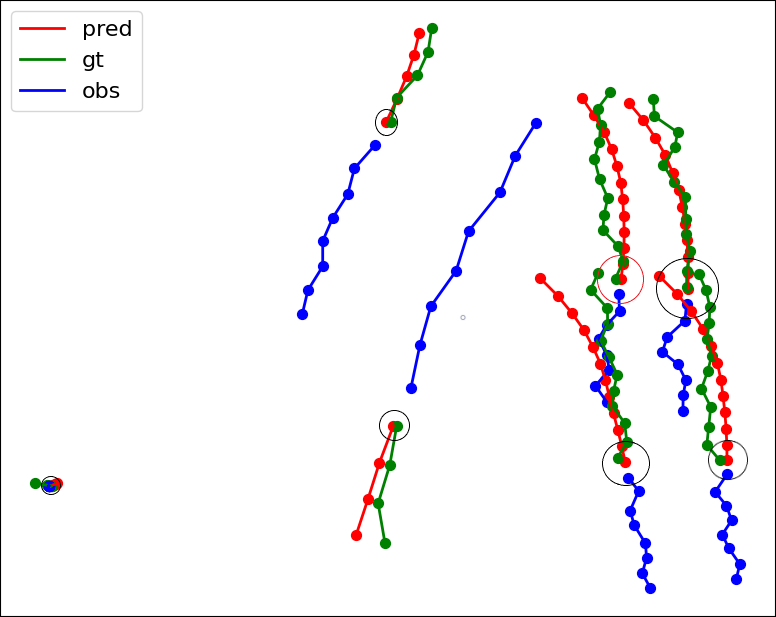}}~~
\subfloat[Temporal Attention Map]{\label{fig:ta}\includegraphics[width=4CM]{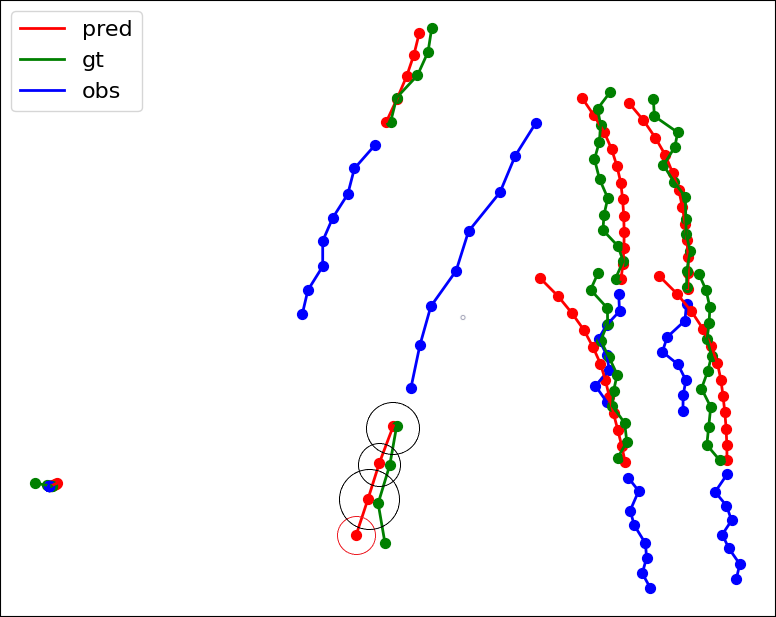}}
\vspace{-5pt}
\subfloat[Crowd Group (scale=3)]{\label{fig:g1}\includegraphics[width=4CM]{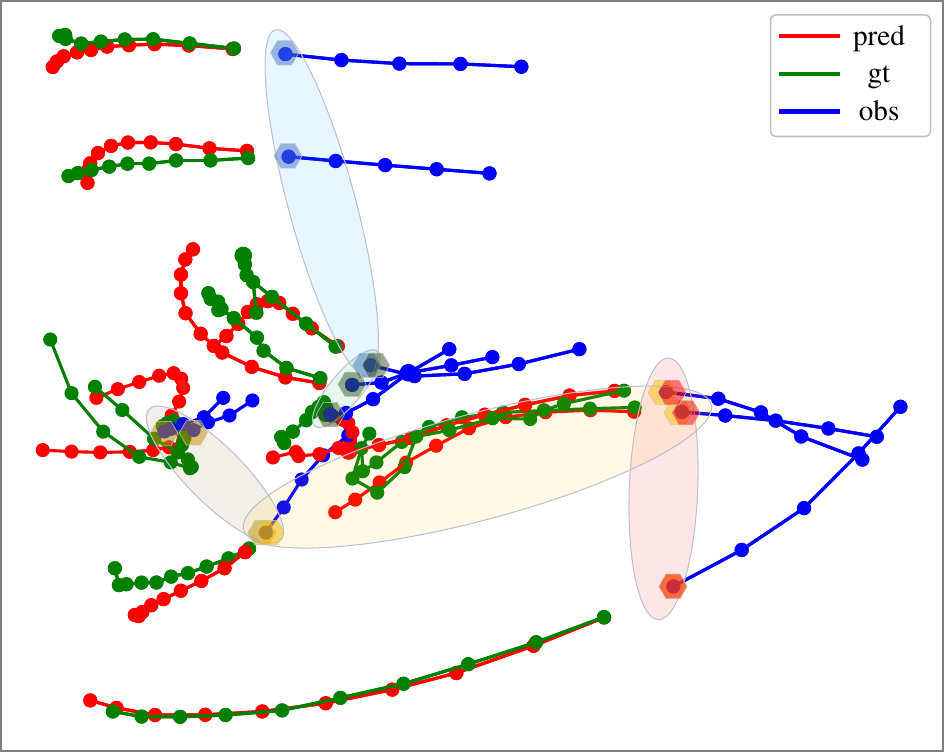}}~~
\subfloat[Crowd Group (Scale=4)]{\label{fig:g2}\includegraphics[width=4CM]{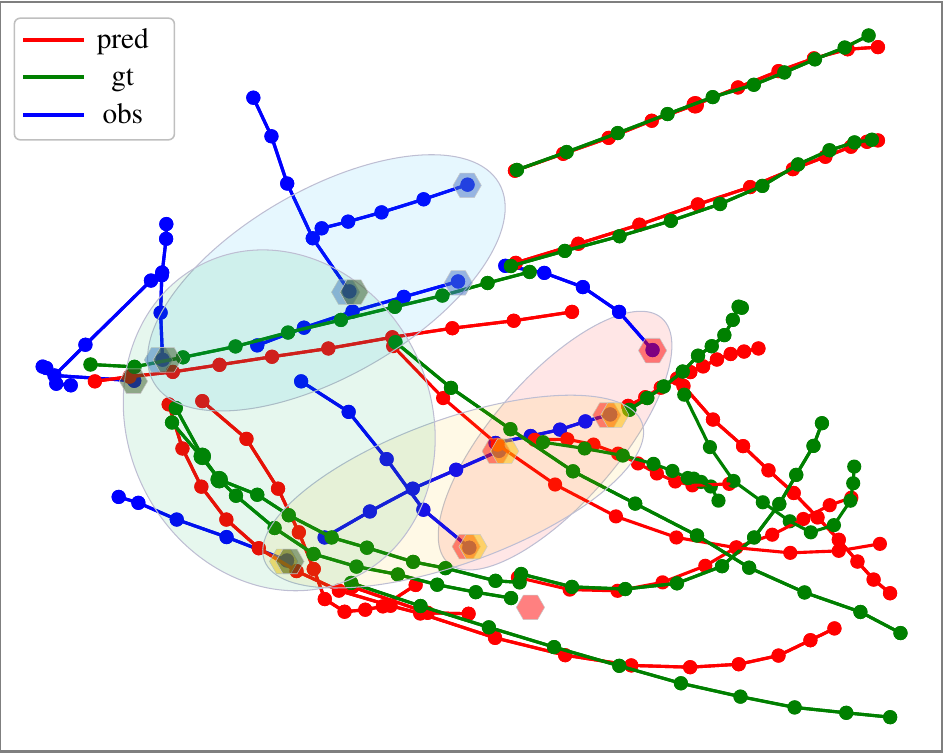}}
\vspace{-0pt}
\caption{The Illustration of groupwise and pairwise interactions in Hyper-STTN: Subfigures (a) and (b) show pairwise attention maps, where red and black circles denote the attention scores of the current agent and neighboring agents, respectively. Subfigures (c) and (d) depict crowd groupwise interactions using hypergraphs at different scales.}
\vspace{-10pt}
\label{fig:attention}
\end{figure}

Fig.~\ref{fig:traj} illustrates the trajectory predictions of Hyper-STTN under both sparse and dense scenarios, alongside two ablation models. In all visualizations, blue dots indicate the observation inputs, while green dots denote the pedestrian ground truth trajectories, allowing clear differentiation from the predicted trajectories. As shown, Hyper-STTN consistently produces highly accurate multi step location forecasts and infers more realistic motion orientations. Fig.~\ref{fig:attention} further highlights how the spatial transformer module enables Hyper-STTN to capture relative spatial interactions among agent pairs, whereas the temporal transformer encodes individual motion attributes. Moreover, in dense scenarios with a growing number of pedestrian groups, inter-group and intra-group crowd correlations exert a stronger influence than in sparse conditions. Consequently, explicitly inferring groupwise interactions yields more accurate long-term forecasts, as evidenced by the trajectories and ADE/FDE metrics achieved by Hyper-STTN and HGNN.



\subsubsection{Effect of Hypergraph Neural Network}

Hyper-STTN consists of two main components: a hypergraph neural network and a spatial-temporal transformer with a fusion module. As shown in Table~\ref{table: Stochastic}, the HGNN branch achieves performance comparable to GroupNet on several datasets, indicating that hypergraph-based interaction modeling is effective, particularly in crowded scenes. This advantage is further reflected in Fig.~\ref{fig:traj}, where the STTN ablation exhibits noticeably larger forecasting errors than HGNN in predicted agent trajectories. These results suggest that the performance gain comes from explicitly modeling higher order groupwise interactions, which are not captured by STTN alone but are effectively introduced by HGNN.


\subsubsection{Effect of Spatial-Temporal Transformer}
Compared with previous transformer-based methods such as STAR, Hyper-STTN improves the average performance from 0.26/0.53 to 0.21/0.33. It also consistently outperforms its STTN and HGNN ablations, reducing the average error from 0.31/0.43 and 0.29/0.48 to 0.21/0.33, respectively. These gains demonstrate the effectiveness of combining transformer-based multi-modal fusion with hypergraph-based group interaction modeling. Moreover, as shown in Fig.~\ref{fig:traj}, Hyper-STTN generates trajectories that are closer to the ground truth than both STTN and HGNN across distinct interaction scenarios.




\vspace{-3pt}
\section{Conclusion}

In this paper, we proposed Hyper-STTN, a hypergraph-based hybrid spatial-temporal transformer for trajectory prediction tasks. Hyper-STTN jointly captures groupwise and pairwise interactions in crowd dynamics through hypergraph and transformer networks. Extensive experiments on the ETH-UCY and NBA datasets demonstrate that Hyper-STTN outperforms existing state-of-the-art prediction algorithms. Future research will focus on enhancing its capability to handle more complex scenarios and on scaling the model for real-time deployments. 




\vspace{-5pt}

\typeout{}
\bibliography{main}
\bibliographystyle{IEEEtran}
\end{document}